\documentclass{article}
\usepackage{ijcai23}

\usepackage{times}
\usepackage{soul}
\usepackage{url}
\usepackage[hidelinks]{hyperref}
\usepackage[utf8]{inputenc}
\usepackage[small]{caption}
\usepackage{graphicx}
\usepackage{amsmath}
\usepackage{amsthm}
\usepackage{booktabs}
\usepackage{algorithm}
\usepackage{algorithmic}
\usepackage[switch]{lineno}

\title{Match and Locate: low-frequency monocular odometry based on deep feature matching}

\author{
Stepan Konev
\and
Yuriy Biktairov
\emails
stevenkonev@gmail.com, \\
biktairo@usc.edu
}

\begin{document}

\maketitle

\begin{abstract}
    Accurate and robust pose estimation plays a crucial role in many robotic systems. Popular algorithms for pose estimation typically rely on high-fidelity and high-frequency signals from various sensors. Inclusion of these sensors makes the system less affordable and much more complicated. In this work we introduce a novel approach for the robotic odometry which only requires a single camera and, importantly, can produce reliable estimates given even extremely low-frequency signal of around one frame per second. The approach is based on matching image features between the consecutive frames of the video stream using deep feature matching models. The resulting coarse estimate is then adjusted by a convolutional neural network, which is also responsible for estimating the scale of the transition, otherwise irretrievable using only the feature matching information. We evaluate the performance of the approach in the AISG–SLA Visual Localisation Challenge and find that while being computationally efficient and easy to implement our method shows competitive results with only around $3^{\circ}$ of orientation estimation error and $2m$ of translation estimation error taking the third place in the challenge.

\end{abstract}

\section{Introduction}

Pose estimation is an essential part of many robotics systems. While some systems may afford carrying many sensors and processing complex algorithms that can utilize incoming data, other systems' resources might be very limited however still requiring reliable solutions. These limitations may include low frequency sensor data along with limitation for number of sensors and their broad availability. The proposed method allows to bypass these limitations for efficient and robust camera pose estimation based on low-frequency RGB images.

Our approach is based on using a two-stage model for predicting the relative pose change between
the each pair of consecutive time frames. Given the relative poses we then compose them in
order to obtain the complete trajectory in the image frame of the first timestamp.

The first stage of the model is a traditional pose estimation technique based on feature
matching. It is capable of accurately estimating the relative rotation of the camera
provided that there are enough objects present in both images simultaneously.

The second stage is a convolutional neural network predicting the adjustment to the first stage estimate. This model is supposed to correct any systematic inaccuracy of the first stage predictor as well as provide an accurate estimate of the relative translation's magnitude, which the first stage model is not able to do.

\section{Problem setting challenges}

Since the experimental setting we analyze in this work is quite specific we first highlight several important details distinguishing it from the traditional monocular odometry.

Affordable cameras we consider may not have a perfectly stable connection to the computing system processing the signal.
Occasionally, the signal might be lost for tens of seconds. When this happens the last image before the interruption and
the first image after might not share any identifiable features. In these situations we cannot hope to estimate the relative
pose change accurately.

These affordable cameras may also not have a perfectly consistent framerate. Meaning that the distribution of the time
intervals between consecutive frames is continuous and might have a considerably heavy tail. This observation motivates
us to pay special attention to the timestamps of the frames instead of assuming a regular time grid.

A part of the image might contain a portion of the actual robotic system the camera is mounted on. This portion's
image would typically be static and would not provide any relevant information about the relative pose change between the frames.
To this extent, we trim any static object present on the image off.

Finally, the lighting conditions may vary a lot between different scenes. The difference might be especially pronounced
when images are taken by an affordable camera with no sophisticated post-processing capabilities. The models we use should
then, ideally, be robust to the perturbations of this sort.

These observations guide the design of our approach and motivate some of the heuristics we use.

\section{Matching-based localization}

\begin{figure}
    \centering
    \includegraphics[width=\columnwidth]{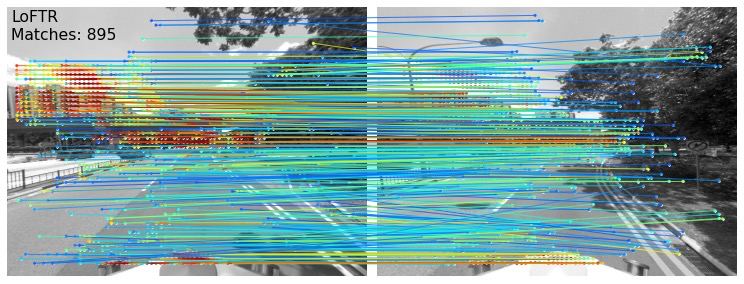}
    \caption{Example of a set of matches between two images.}
    \label{fig:matches}
\end{figure}

At the heart of our approach is the classic technique of reconstructing the relative pose
between two camera images based on a set of matched points
$M = \left\{ (x_i, y_i) \right\}$, where $x_i \in I_1$ and $y_i \in I_2$.
Each match $(x, y)$ indicates that point $x$ in the first image $I_1$ and point $y$
in the second image $I_2$ both correspond to the same original point in the 3d space. Figure
\ref{fig:matches} illustrates the idea by connecting the matched points with a colored line.
Given an accurate enough set of at least 8 matches we can reconstruct the relative pose
between the cameras up to a scale factor. The scale factor cannot be determined based solely
on camera images: intuitively, images only contain the information regarding the directions
of rays coming to the camera, but these directions stay the same when we scale the world up
or down.

\subsection{Deep feature matching models}

Traditionally, so-called sparse feature matching is used for this kind application. The most
prominent example of such application would be the monocular visual SLAM, frequently used as
a localisation solution for drones and in-door robots. Sparsity implies that only select
points from each image are attempted to be matched against each other. The selection
criterion is usually based on a kind of edge detection algorithm and the matching relies on
comparing a set of features extracted from around the candidate points.

These techniques work quite good for the high framerate (or, alternatively, slow motion)
situations, where the relative pose between the consecutive frames is not far from the
identity pose. This is not the case, however, in our setting where the typical time
interval between the frames is around $1s$ and the system's speed is on the
scale of $10 m/s$. Sometimes the relative pose comprises such a big rotation that
the consecutive images do not have any semantic intersection. For this reason we use dense
feature matching instead.

Dense feature matching algorithms do not rely on finding distinct boundaries between objects
and produce higher number of matches along with the confidence scores for each match.
Recently, a number of deep-learning-based models for dense feature matching were published.
According to the evaluation results these are capable of producing decent matches even for
drastically different camera poses. We experimented with LoFTR\cite{sun2021loftr},
DKM \cite{edstedt2022dkm} and CoTracker\cite{karaev2023cotracker} feature matching models
and found the former to be better suited for our task.

\subsection{Pose reconstruction}

The go-to approach to the matching-based pose reconstruction is the RANSAC algorithm\cite{fischler1981random} capable of handling a large number of erroneous matches, or a more recent development of the algorithm – GC-RANSAC\cite{barath2018graph}.

During our evaluation we found that GC-RANSAC does not provide noticeable improvement
over the conventional RANSACK, so we used the latter for our main model. Another set of
evaluations shows that the optimal values of parameters for our case are
\begin{align*}
prob &= 0.99999, \\
threshold &= 0.9.
\end{align*}

\subsection{Coarse translation estimation}

Since the scale of the relative pose cannot be determined by the images alone, we use the
following simple heuristic for our base model: we assume that the absolute value of the
translation is the same for every pair of consecutive frames in the trajectory. This
simplification is inaccurate since there are both stationary sections and long temporal
discontinuities present in the train trajectories. It is, however, a decent base for the
refinement model to build upon.

We choose this constant translation's magnitude to be $10m$ based on running the approach on the
training set in our experimental evaluation.

\subsection{Constant rotation heuristic}

Another important heuristic we use covers the pairs of frames where no meaningful semantic
intersection is present between the images. We use the number of matches with high enough
confidence scores as an indicator that such a situation is encountered.

Similarly to the translation heuristic, here we assume a constant rotation. We notice that
the movement of the camera is generally flat, so we assume a rotation around the vertical
axis which almost coincides with $x$ axis of the image frame. As to the value of this
rotation – we choose $\pi$ as in many cases such a situation can be attributed to the
system making a u-turn or a similar maneuver.

\section{CNN refinement model}

Given a matching-based backbone model, we train a convolutional neural network to predict
adjustments to the estimate provided by the backbone. Introducing this second stage to the
pipeline can potentially lead to the following improvements:
\begin{itemize}
    \item CNN can provide sensible non-constant estimates of the translation;
    \item CNN can adjust the prior rotation based on the actual statistics of the dataset;
    \item CNN can compensate for any systematic bias in the matching-based model.
\end{itemize}

\subsection{Additional features}

In order to facilitate the training of the refinement model we preprocess the dataset by
adding three feature maps:
\begin{enumerate}
    \item the monocular depth estimate of the first frame extracted by MiDaS\cite{ranftl2020towards};
    \item the monocular depth estimate of the second frame extracted by MiDaS;
    \item the optical flow estimate extracted by RAFT\cite{teed2020raft};
    \item time interval encoding;
\end{enumerate}
These feature maps are only extracted once. We do not train the corresponding models and do not extract the features again after augmenting the RGB images during traning.

\subsection{Architecture}

We experimented with several pre-trained convolutional vision models from timm\cite{rw2019timm} including versions of EfficientNet\cite{tan2019efficientnet} and ResNet\cite{he2016deep},

\subsection{Pose representation}

For the training purposes we represent the relative pose as a pair of $3d$ translation
vector and $4d$ unit quaternion corresponding to the rotation. We model the predicted
translation as the sum of the matching-based estimate and the output of the final linear
layer of the CNN
\[
t = t_{base} + t_{CNN}.
\]
We model the predicted rotation in a similar way, normalizing the result to preserve the
unitary length of the quaternion
\[
q = \frac{q_{base} + q_{CNN}}{|q_{base} + q_{CNN}|}.
\]

In some evaluations we also eliminate the ambiguity of representing a rotation with
a quaternion by enforcing $w > 0$. We do it by transforming the real part of the quaternion
prior to normalization
\[
w = \exp (w_{prelim}).
\]
While it does eliminate the ambiguity and should, therefore, enhance the generalization, it
also makes representation the rotations of magnitude $\sim \pi$ discontinuous. It make the
performance of such standardisation highly dependent on the structure of the dataset. We
couldn't make a decisive conclusion in this regard.

The loss function we use for training directly mimics the metrics used for the evaluation
\begin{align*}
L_r &= 2 \arccos(|q \cdot q_{GT}|), \\
L_t &= |t - t_{GT}|_2,
\end{align*}
where we linearly extrapolate $\arccos$ in the interval $[1 - \varepsilon, 1]$ in order to
avoid infinite gradients.

\subsection{Augmentations}

As the training dataset is quite limited we use a vast range of augmentations in order to
facilitate the generalization of the CNN model.

We apply a big suite of standard visual augmentations supported by Albumentations\cite{buslaev2020albumentations} to the
model so it can more easily generalize to new lighting conditions.

We use the approximate planarity of motion again and introduce the vertical reflection of the images as an augmentation. This transformation does not really preserve the distribution over the images since the left-hand-side roads become right-hand-side, but these high-level changes should not affect the basic localisation task too much. Reflection of the images should also be accompanied with the corresponding amendments to the optical flow map and the target relative pose.

Finally, we augment the dataset by applying a perspective transformation corresponding to small additional rotation of the camera to the second image. We then crop the image to hide the border, where we might not have the image values after the transformation. This transformation also requires adjusting the target relative pose and optical flow map.

\section{Inference time}

Running on a machine with an nvidia rtx 3090 GPU our approach takes around $\sim 1$s to process a single pair of consecutive images. With the typical framerate of the trajectories in the dataset also being close to 1s we are confident that with a couple optimizations the approach may be robustly run in real time.

\section{Evaluation}

We evaluate our approach in the AISG–SLA Visual Localisation Challenge. The results of this evaluation
are displayed in table \ref{table:evaluation}.

The upper four models do not use matching-based coarse estimates as a baseline. They only contain stage 2
of our approach: a CNN predicting the $6d$ pose. The results indicate that applying augmentations we
designed for the training is beneficial for both a single CNN model and an ensemble of models.
The results also suggest that an ensemble of CNNs generalizes better to the private testing dataset,
which is a known effect of ensembling.

Finally, the last entry in table \ref{table:evaluation} show the performance of our complete model
with both stages enabled. We can clearly see that the first matching-base stage of the approach
is indeed very important and improves the results significantly.

\begin{table}[t]
\caption{Experimental evaluation.}\label{table:evaluation}
\centering
\begin{tabular}{l|lll}
Model & $R_r^{pub.}$ & $R_r^{priv.}$ & $R_t^{priv.},m$ \\
\hline
2\textsuperscript{nd} stage only & 0.041 & 0.094 & 12.0 \\
2\textsuperscript{nd} stage only + aug. & 0.039 & 0.072 & 11.7 \\
2\textsuperscript{nd} stage ensemble & 0.043 & 0.069 & 10.8 \\
2\textsuperscript{nd} stage ensemble + aug. & 0.041 & 0.064 & 11.0 \\
both stages & \textbf{0.032} & \textbf{0.046} & \textbf{2.0} \\

\hline
\end{tabular}
\end{table}

\bibliographystyle{named}
\bibliography{ijcai23}

\begin{thebibliography}{}

\bibitem[\protect\citeauthoryear{Barath and Matas}{2018}]{barath2018graph}
Daniel Barath and Ji{\v{r}}{\'\i} Matas.
\newblock Graph-cut ransac.
\newblock In {\em Proceedings of the IEEE conference on computer vision and
  pattern recognition}, pages 6733--6741, 2018.

\bibitem[\protect\citeauthoryear{Buslaev \bgroup \em et al.\egroup
  }{2020}]{buslaev2020albumentations}
Alexander Buslaev, Vladimir~I Iglovikov, Eugene Khvedchenya, Alex Parinov,
  Mikhail Druzhinin, and Alexandr~A Kalinin.
\newblock Albumentations: fast and flexible image augmentations.
\newblock {\em Information}, 11(2):125, 2020.

\bibitem[\protect\citeauthoryear{Edstedt \bgroup \em et al.\egroup
  }{2022}]{edstedt2022dkm}
Johan Edstedt, Ioannis Athanasiadis, Mårten Wadenbäck, and Michael Felsberg.
\newblock Dkm: Dense kernelized feature matching for geometry estimation, 2022.

\bibitem[\protect\citeauthoryear{Fischler and
  Bolles}{1981}]{fischler1981random}
Martin~A Fischler and Robert~C Bolles.
\newblock Random sample consensus: a paradigm for model fitting with
  applications to image analysis and automated cartography.
\newblock {\em Communications of the ACM}, 24(6):381--395, 1981.

\bibitem[\protect\citeauthoryear{He \bgroup \em et al.\egroup
  }{2016}]{he2016deep}
Kaiming He, Xiangyu Zhang, Shaoqing Ren, and Jian Sun.
\newblock Deep residual learning for image recognition.
\newblock In {\em Proceedings of the IEEE conference on computer vision and
  pattern recognition}, pages 770--778, 2016.

\bibitem[\protect\citeauthoryear{Karaev \bgroup \em et al.\egroup
  }{2023}]{karaev2023cotracker}
Nikita Karaev, Ignacio Rocco, Benjamin Graham, Natalia Neverova, Andrea
  Vedaldi, and Christian Rupprecht.
\newblock Cotracker: It is better to track together.
\newblock {\em arXiv preprint arXiv:2307.07635}, 2023.

\bibitem[\protect\citeauthoryear{Ranftl \bgroup \em et al.\egroup
  }{2020}]{ranftl2020towards}
Ren{\'e} Ranftl, Katrin Lasinger, David Hafner, Konrad Schindler, and Vladlen
  Koltun.
\newblock Towards robust monocular depth estimation: Mixing datasets for
  zero-shot cross-dataset transfer.
\newblock {\em IEEE transactions on pattern analysis and machine intelligence},
  44(3):1623--1637, 2020.

\bibitem[\protect\citeauthoryear{Sun \bgroup \em et al.\egroup
  }{2021}]{sun2021loftr}
Jiaming Sun, Zehong Shen, Yuang Wang, Hujun Bao, and Xiaowei Zhou.
\newblock Loftr: Detector-free local feature matching with transformers.
\newblock In {\em Proceedings of the IEEE/CVF conference on computer vision and
  pattern recognition}, pages 8922--8931, 2021.

\bibitem[\protect\citeauthoryear{Tan and Le}{2019}]{tan2019efficientnet}
Mingxing Tan and Quoc Le.
\newblock Efficientnet: Rethinking model scaling for convolutional neural
  networks.
\newblock In {\em International conference on machine learning}, pages
  6105--6114. PMLR, 2019.

\bibitem[\protect\citeauthoryear{Teed and Deng}{2020}]{teed2020raft}
Zachary Teed and Jia Deng.
\newblock Raft: Recurrent all-pairs field transforms for optical flow.
\newblock In {\em Computer Vision--ECCV 2020: 16th European Conference,
  Glasgow, UK, August 23--28, 2020, Proceedings, Part II 16}, pages 402--419.
  Springer, 2020.

\bibitem[\protect\citeauthoryear{Wightman}{2019}]{rw2019timm}
Ross Wightman.
\newblock Pytorch image models.
\newblock \url{https://github.com/rwightman/pytorch-image-models}, 2019.

\end{thebibliography}

\end{document}